\documentclass[10pt,twocolumn,letterpaper]{article}
\PassOptionsToPackage{dvipsnames,table}{xcolor}

\usepackage{cvpr}
\usepackage[dvipsnames]{xcolor}

\definecolor{cvprblue}{rgb}{0.21,0.49,0.74}

\usepackage{times}
\usepackage{graphicx}
\usepackage{amsmath}
\usepackage{amssymb}
\usepackage{booktabs}
\usepackage{rotating}
\usepackage{paralist}
\usepackage{caption}
\usepackage{subcaption}
\usepackage[linesnumbered,boxed]{algorithm2e}

\usepackage{makecell}

\usepackage{times}
\usepackage{array}
\usepackage{multirow}
\usepackage{ragged2e}
\usepackage{booktabs}
\usepackage{wrapfig}

\usepackage{graphicx}
\usepackage{amsmath}
\usepackage{amssymb}
\usepackage[pagebackref=true,breaklinks=true,colorlinks,bookmarks=false,citecolor=blue,linkcolor=blue]{hyperref}
\usepackage{makecell}
\usepackage[accsupp]{axessibility}
\usepackage{color, colortbl}
\definecolor{LightCyan}{rgb}{0.88,1,1}

\hypersetup{
    colorlinks=true,
    urlcolor=magenta,
    citecolor=blue,
}

\makeatletter\if@twocolumn\PassOptionsToPackage{switch}{lineno}\else\fi\makeatother

\usepackage[T1]{fontenc}
\usepackage{multirow}
\usepackage{booktabs}
\usepackage{comment}
\usepackage{color}

\usepackage{textcomp}
\usepackage{siunitx}
\usepackage{tabulary}
\usepackage{makecell}
\usepackage{multirow}
\usepackage{booktabs}
\usepackage{hyperref}

\makeatletter
\newcommand\figref{Figure~\ref}
\newcommand{\tabref}[1]{Table~\ref{#1}}
\newcolumntype{P}[1]{>{\centering\arraybackslash}p{#1}}
\newcolumntype{M}[1]{>{\centering\arraybackslash}m{#1}}
\let\ts@includegraphics\includegraphics

\usepackage{float}

\usepackage[capitalize]{cleveref}
\crefname{section}{Sec.}{Secs.}
\Crefname{section}{Section}{Sections}
\Crefname{table}{Table}{Tables}
\crefname{table}{Tab.}{Tabs.}

\def\paperID{13711} 
\def\confName{CVPR}
\def\confYear{2024}

\begin{document}

\title{DSL-FIQA: Assessing Facial Image Quality via Dual-Set Degradation Learning and Landmark-Guided Transformer}

\author{
Wei-Ting Chen\textsuperscript{1,2†}
\quad Gurunandan Krishnan\textsuperscript{2}
\quad Qiang Gao\textsuperscript{2}
\quad Sy-Yen Kuo\textsuperscript{1} 
\quad Sizhuo Ma\textsuperscript{2*}
\quad Jian Wang\textsuperscript{2*$\blacklozenge$}
\\\\
\hspace{-8mm}\textsuperscript{1}National Taiwan University\quad \textsuperscript{2}Snap Inc.
}

\maketitle

\begin{abstract}
Generic Face Image Quality Assessment (GFIQA) evaluates the perceptual quality of facial images, which is crucial in improving image restoration algorithms and selecting high-quality face images for downstream tasks.
We present a novel transformer-based method for GFIQA, which is aided by two unique mechanisms. 
First, a ``\textbf{D}ual-\textbf{S}et Degradation Representation \textbf{L}earning'' (DSL) mechanism uses facial images with both synthetic and real degradations to decouple degradation from content, ensuring generalizability to real-world scenarios. This self-supervised method learns degradation features on a global scale, providing a robust alternative to conventional methods that use local patch information in degradation learning. 
Second, our transformer leverages facial landmarks to emphasize visually salient parts of a face image in evaluating its perceptual quality.
We also introduce a balanced and diverse Comprehensive Generic Face IQA (CGFIQA-40k) dataset of 40K images carefully designed to overcome the biases, in particular the imbalances in skin tone and gender representation, in existing datasets.
Extensive analysis and evaluation demonstrate the robustness of our method, marking a significant improvement over prior methods.
\end{abstract}

\newcommand\blfootnote[1]{%
\begingroup
\renewcommand\thefootnote{}\footnote{#1}%
\addtocounter{footnote}{-1}%
\endgroup
}
\blfootnote{Project Page: \href{https://dsl-fiqa.github.io}{https://dsl-fiqa.github.io}}
\blfootnote{† Part of the work done during internship at Snap Research.}
\blfootnote{* Co-corresponding authors}
\blfootnote{$\blacklozenge$ Project lead}


\begin{figure}[t!]
  \centering
  \begin{subfigure}[b]{0.49\columnwidth} 
    \includegraphics[width=\textwidth]{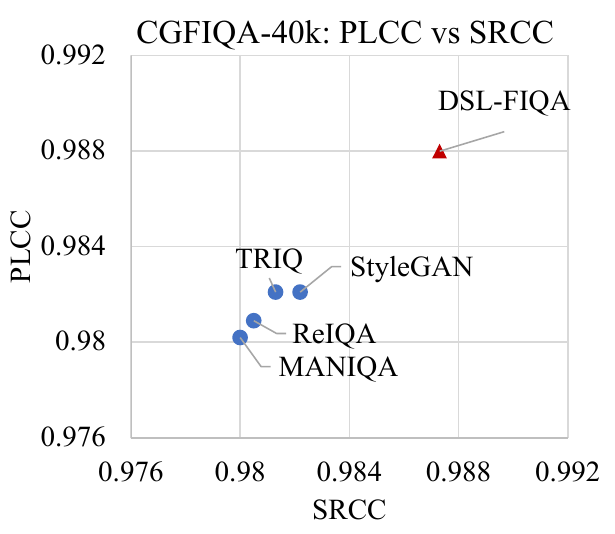} 
    \caption{CGFIQA-40k}
    \label{fig:subim1}
  \end{subfigure}
  \begin{subfigure}[b]{0.49\columnwidth} 
    \includegraphics[width=\textwidth]{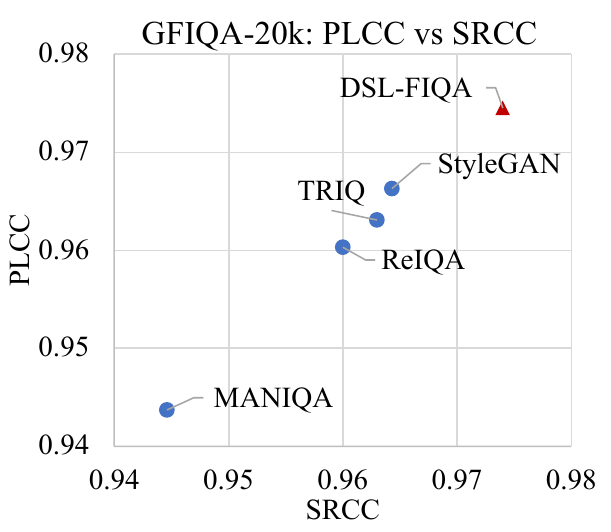} 
    \caption{GFIQA-20k}
    \label{fig:subim2}
  \end{subfigure}
\caption{\textbf{PLCC vs. SRCC Comparison on CGFIQA-40k and GFIQA-20k~\cite{su2023going} datasets.} DSL-FIQA, denoted by red triangular points, outperforms other methods (ReIQA~\cite{saha2023re}, StyleGAN-IQA~\cite{su2023going}, MANIQA~\cite{yang2022maniqa}, TRIQ~\cite{tu2021rapique}) and can provide a superior image quality assessment of facial images.} 

  \label{fig:teasor}
\end{figure}

\section{Introduction}
In the digital era, face images hold a central role in our visual experiences, necessitating a robust metric for assessing their perceptual quality. This metric is crucial for not only evaluating and improving the performance of face restoration algorithms but also for assuring the quality of training datasets for generative models~\cite{karras2017progressive,ruiz2023dreambooth}. Designing an effective metric for face image quality assessment presents significant challenges. The inherent complexity of human faces, characterized by nuanced visual features and expressions, greatly impacts perceived quality~\cite{su2023going}. Additionally, obtaining subjective scores such as Mean Opinion Scores (MOS) is difficult due to the limited availability of licensed face images and the inherent ambiguity in subjective evaluations. Compounding these challenges are facial occlusions caused by masks and accessories, which add another layer of complexity to the assessment process.

Decades of research on image quality assessment (IQA) on general images~\cite{DBLP:journals/corr/abs-2108-05997, yang2022maniqa,DBLP:journals/corr/abs-2110-13266,zhang2023blind,gao2022image}, or general IQA (GIQA), has demonstrated reliable performance across various generic IQA datasets~\cite{sheikh2006statistical,ponomarenko2009tid2008,lin2019kadid,hosu2020koniq,ghadiyaram2015massive,ying2020patches}. However, when such methods are applied to faces, they often overlook the distinct features and subtleties inherent to faces,
making them less effective for face images.

Another thread of research focuses on biometric face quality assessment (BFIQA)~\cite{best2018learning,boutros2023cr,hernandez2019faceqnet,lijun2019multi,ou2021sdd,rose2019deep,schlett2022face,yang2019dfqa}, where the goal is to ensure the quality of a given face image for robust biometric recognition. While recognizability is achieved by including factors unique to faces like clarity, pose, and lighting, 
it does not guarantee accurate assessment of \emph{perceptual} degradation.

A significant stride forward is made by~\cite{su2023going}, which clearly defines the problem of generic face IQA (GFIQA): GFIQA focuses exclusively on the perceptual quality of face images, as opposed to BFIQA. Their approach leverages pre-trained generative models (StyleGAN2~\cite{karras2020analyzing}) to extract latent codes from input images, which are then used as references for quality assessment. Although their method shows promising prediction performance, its effectiveness reduces when input images deviate significantly in shooting angles~\cite{su2023going} or quality~\cite{poirier2023robust} from the StyleGAN2 training data, limiting its applicability and accuracy to real-world scenarios.

In this paper, we tackle the challenge of GFIQA by proposing a transformer-based method specifically designed to address the limitations of the aforementioned methods. Inspired by modern GIQA techniques~\cite{saha2023re}, we propose a degradation extraction module that obtains degradation representations from input images as intermediate features to aid the regression of quality scores, which is pre-trained via self-supervised learning. However, the existing degradation representation learning scheme \cite{saha2023re} often makes an oversimplified assumption that the degradation is uniform across different patches of an image while being distinct from those of other images. This assumption does not hold for real-world data, where diverse degradations within a single image exist due to variations in lighting, motion, camera focus and so on. These inconsistencies may impair the effectiveness of degradation extraction and subsequently hinder the accuracy of quality score prediction.

To this end, we introduce a strategy termed ``\textbf{D}ual-\textbf{S}et Degradation Representation \textbf{L}earning'' (DSL), which breaks the limits of traditional patch-based learning and extracts degradation representations from a global perspective in degradation learning. This approach is enabled by establishing correspondences between a \emph{controlled} dataset of face images with \emph{synthetic} degradations and a comprehensive \emph{in-the-wild} dataset with \emph{realistic} degradations, offering a comprehensive framework for degradation learning. This degradation representation is injected into the transformer decoder~\cite{dosovitskiy2020image} via cross-attention~\cite{rombach2022high}, enhancing the overall sensitivity to various kinds of challenging real-world image degradations.

Furthermore, inspired by \cite{wang2022survey}, we utilize the strong correlation between facial image quality and salient facial components such as mouth and eyes. We incorporate landmark detection to localize and feed them as input to our model. This extra module allows our model to autonomously learn to focus on these critical facial components and understand their correlation with the perceptual quality of faces, 
which helps predict a regional confidence map that aggregates local quality evaluations across the entire face.

Existing datasets such as GFIQA-20k~\cite{su2023going} and PIQ23~\cite{Chahine_2023_CVPR} suffer from limited size or unbalanced distribution. To bridge this gap, we introduce the Comprehensive Generic Face IQA Dataset (CGFIQA-40k), which comprises 40K images with more balanced diversity in \emph{gender} and \emph{skin tone}. We also include face images with facial occlusions. We believe this dataset will be a valuable resource to fuel and inspire future research.

To summarize, our contributions are as follows:
\begin{compactitem}
\item We design a transformer-based method specifically designed for GFIQA, predicting perceptual scores for face images.

\item We propose ``Dual-set Degradation Representation Learning'' (DSL), a self-supervised approach for learning degradation features globally. This method effectively captures global degradation representations from both synthetically and naturally degraded images, enhancing the learning process of degradation characteristics.

\item We enhance our model's attention to salient facial components by integrating facial landmark detection, enabling a holistic quality evaluation that adaptively aggregates local quality assessment across the face.

\item We construct the Comprehensive Generic Face IQA Dataset (CGFIQA-40k), a collection of 40K face images designed to offer comprehensive and balanced representation in terms of gender and skin tone.

\end{compactitem}

 \section{Related Work}
\subsection{Quality Assessment of Face Images}
Recent work on Face Image Quality Assessment (FIQA) can be categorized into two major branches~\cite{wang2022survey}: BFIQA and GFIQA. 
BFIQA originates from biometric studies, focusing on the quality of face images for recognition systems. On the other hand, GFIQA encompasses a wider scope, concentrating on the perceptual degradation of image quality.

\noindent\textbf{Biometric Face Image Quality Assessment (BFIQA)}:
BFIQA evaluates the quality of face images for biometric applications such as face recognition, which often assess images based on established standards \cite{grother2020ongoing,schlett2021face,sellahewa2010image}. Recent progress in the field has been around learning-based strategies, assessing quality via performance of a recognition system~\cite{boutros2023cr,meng2021magface,terhorst2020ser,bharadwaj2013can,chen2014face}. Some studies have adopted manual labeling, using a set of predefined characteristics as binary constraints \cite{zhao2019face}, while others have investigated subjective aspects of image quality \cite{best2018learning}. However, adopting BFIQA methods does not give the best performance when the emphasis is on perceptual quality, which will be demonstrated in the results section.

\begin{figure*}[t!]
\centering \includegraphics[width=0.79\textwidth,page=1,]{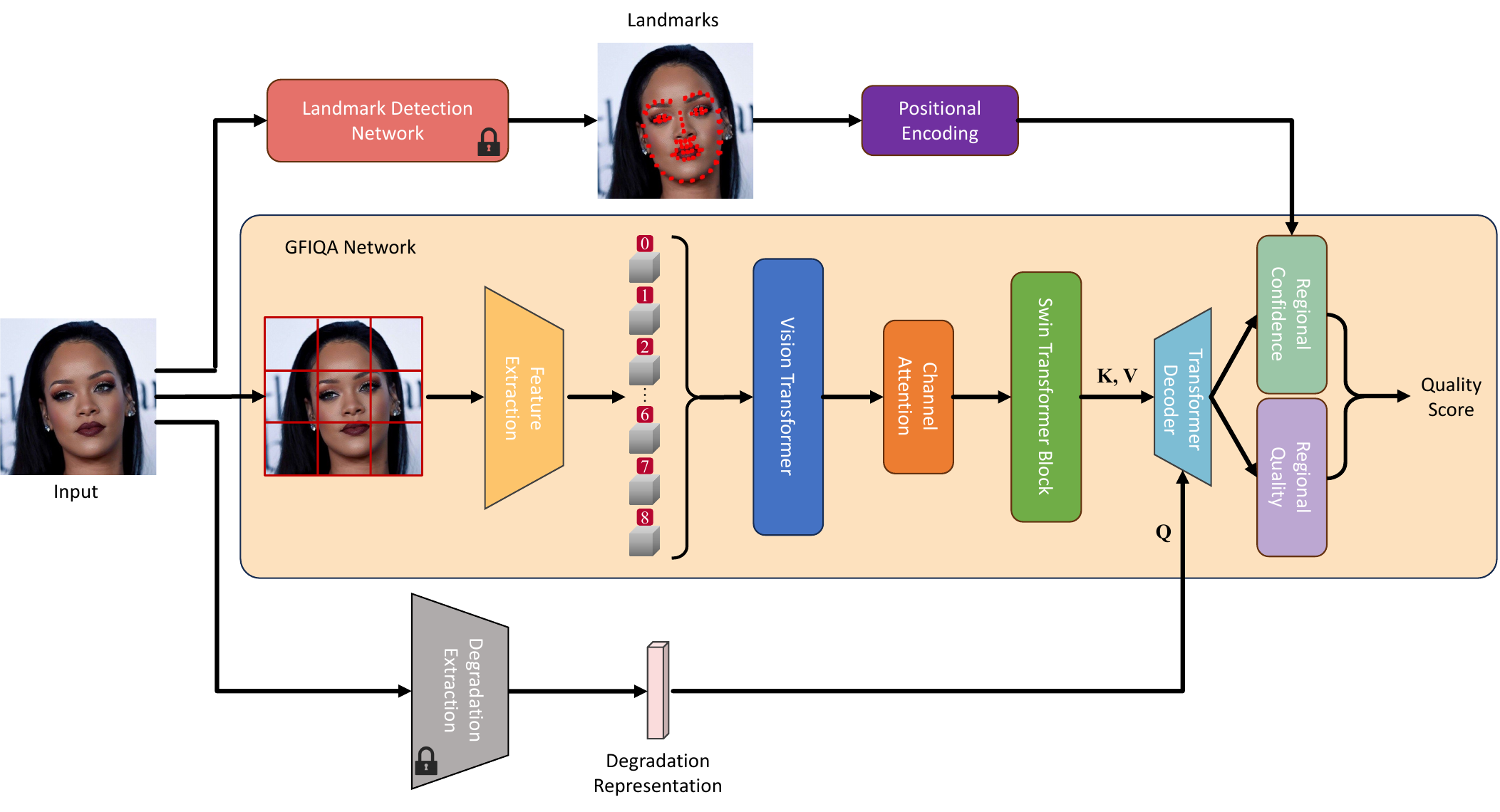}{}
\makeatother 
\caption{\textbf{Overview of our proposed model.} The model contains a core GFIQA network, a degradation extraction network, and a landmark detection network. In our approach, face images are cropped into several patches to fit the input size requirements of the pre-trained ViT feature extractor (See \cref{sec:model_overview}). Each patch is then processed individually, and their Mean Opinion Scores (MOS) are averaged to determine the final quality score. For clarity in the figure, the segmentation of the image into patches is not shown.}
\label{fig:overview}
\end{figure*}

\noindent\textbf{Generic Face Image Quality Assessment (GFIQA)}:
GFIQA is a recently defined task \cite{wang2022survey,su2023going}, which prioritizes perceptual degradation in face images instead. Initiatives like Chahine \textit{et al.} \cite{Chahine_2023_CVPR} highlight the relevance of social media-driven portrait photography, though their PIQ23 dataset remains restricted in size. Su \textit{et al.} \cite{su2023going} bridges this gap with the larger GFIQA-20k dataset, but it falls short in diversity, lacking in long-tail samples and balanced representation. Their generative prior-based method, while effective, struggles with non-standard images, showing limitations of StyleGAN2-dependent models~\cite{karras2019style}. This demonstrates the necessity for a more comprehensive dataset and robust GFIQA solution.

\subsection{General Image Quality Assessment}
Traditional General Image Quality Assessment (GIQA)~\cite{gao2022image,chen2022teacher} methods like BRISQUE~\cite{mittal2012no}, NIQE~\cite{mittal2012making}, and DIIVINE~\cite{moorthy2011blind} are built upon traditional statistical models, which work decently on datasets with constrained size but have faced limitations with complex real-world images. The advent of deep learning has given rise to groundbreaking GIQA methods.
RAPIQUE~\cite{tu2021rapique} and DB-CNN~\cite{zhang2018blind} set new standards in adaptability and accuracy. A further innovation was seen in transformer-based models, including MUSIQ~\cite{ke2021musiq} and MANIQA~\cite{yang2022maniqa}, with significantly improved prediction precision. The domain was expanded by CONTRIQUE's~\cite{DBLP:journals/corr/abs-2110-13266} self-supervised paradigm and Zhang \textit{et al.}'s~\cite{zhang2023blind} vision-language multitask learning. Saha \textit{et al.}~\cite{saha2023re} uniquely integrated low and high-level features in an unsupervised manner, emphasizing perceptually relevant quality features.

However, the applicability of these advancements to face images is debatable, as they often overlook facial features specifically critical for perceptual quality, suggesting a gap ripe for exploration with face-centric quality assessment models.

\begin{figure*}[t!]
\centering \includegraphics[width=0.85\textwidth,page=1]{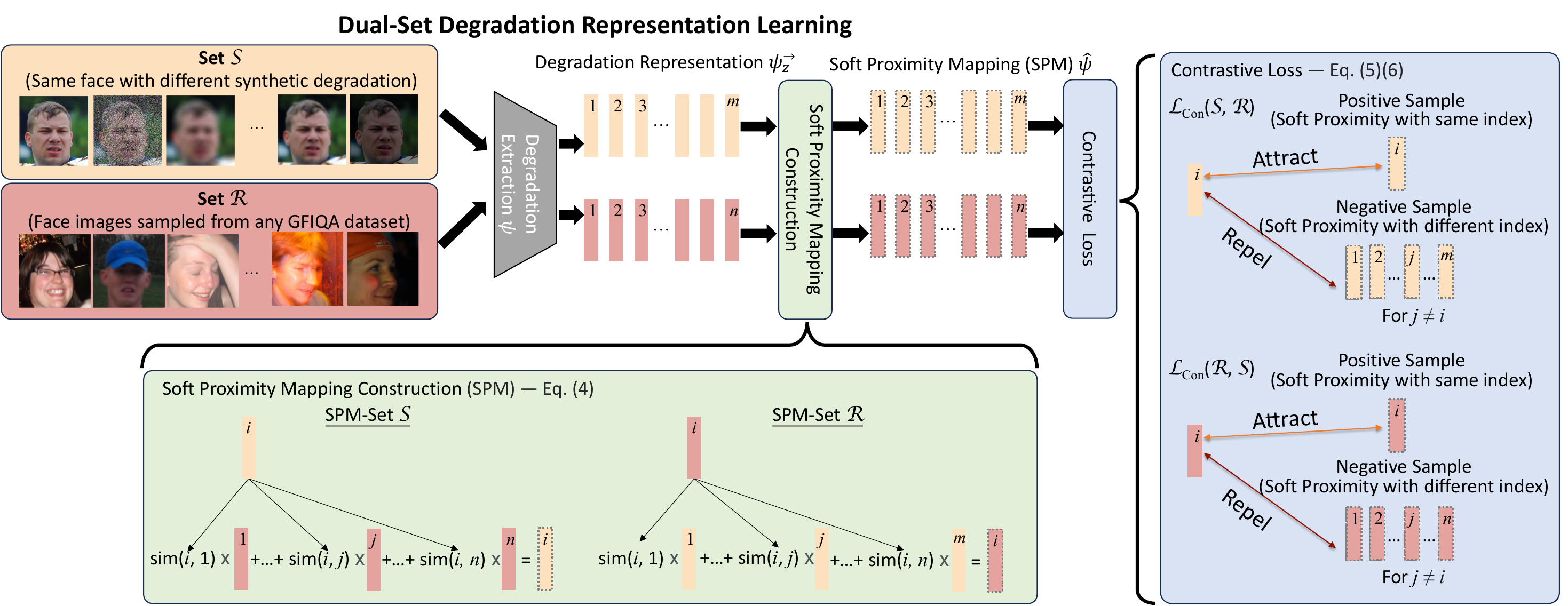}{}
\makeatother 
\caption{\textbf{Dual-Set Degradation Representation Learning (DSL) Illustrated.} On the left, the process of contrastive optimization is depicted, utilizing two unique image sets. Degradation representations are extracted, followed by soft proximity mapping (SPM) calculations and contrastive optimization, compelling the degradation encoder to focus on learning specific degradation features. The right side emphasizes the bidirectional characteristic of our approach, highlighting the comprehensive strategy for identifying and understanding image degradations through contrastive learning.} 
\label{fig:dce}
\end{figure*}

\section{Proposed Method} \label{sec:method}
\subsection{Model Overview} \label{sec:model_overview}
Figure \ref{fig:overview} illustrates our method: Given an input image $\boldsymbol{I} \in \mathbb{R}^{H \times W \times 3}$, our model estimates its perceptual quality score. In the following, we briefly summarize its components.

\noindent\textbf{Feature Extraction and Refinement:} The image initially undergoes feature extraction~\cite{yang2022maniqa} via a pre-trained Vision Transformer (ViT)~\cite{dosovitskiy2020image}, followed by a Channel Attention Block~\cite{hu2018squeeze} that emphasizes relevant inter-channel dependencies. Subsequently, a Swin Transformer Block~\cite{liu2021swin} refines these features, capturing subtle image details.

\noindent\textbf{Degradation Extraction:} In parallel, a dedicated module identifies and isolates perceptual degradations within the image, providing a nuanced representation of image quality degradations.

\noindent\textbf{Feature Integration and Quality Estimation:} The degradation features, once extracted, are integrated with the outputs from the Swin Transformer within a transformer decoder. This integration employs cross-attention, a technique inspired by Stable Diffusion~\cite{rombach2022high}, to enhance the model's sensitivity to degradation. The combined features are then directed into two MLP branches. The first branch predicts the \emph{regional confidence}, while the second estimates the \emph{regional quality score}. Finally, these outputs are combined through a weighted sum to determine the overall quality score of the image.

\noindent\textbf{Landmark-Guided Mechanism:} A landmark detection network identifies facial key points, influencing the regional confidence evaluation and ensuring that essential facial features improve the final quality score.

During the training of the core GFIQA network, the landmark and degradation modules remain fixed, leveraging their pre-trained knowledge. Notably, we avoid resizing input images to fit the fixed input dimensions of the pre-trained ViT, which could distort quality predictions \cite{ke2021musiq}. Instead, we crop the image, process each part independently, and average the resulting MOS predictions for a consolidated image quality score. This approach maintains the original dimensions of the image and, consequently, the correctness of perceptual quality assessment.

In the following subsections, we highlight the main technical contribution of our model design: our degradation extraction module and landmark-guided mechanism.

\subsection{Self-Supervised Dual-Set Degradation Representation Learning}
Before presenting our proposed approach, we provide a concise overview of the existing patch-based degradation learning methods.
\subsubsection{Patch-based Degradation Learning} \label{sec:patch_based}
Existing degradation extraction methods (\cite{saha2023re,zhao2023quality,huang2023counting}) assume that patches from the same image share similar degradation for contrastive learning. In this framework, patches extracted from the same image are positive samples, while those from different images are negative samples. The patches are encoded into degradation representations ($x$, $x^+$, and $x^-$) for the query, positive, and negative samples. The contrastive loss function is designed to enhance the similarity between $x$ and $x^+$ and dissimilarity between $x$ and $x^-$, which is given by:
\begin{equation}
\scalebox{1.0}{$
\mathcal{L}_{Patch}(x, x^+, x^-) = -\log \frac{\exp(x \cdot x^+ / \theta)}{\sum_{n=1}^{N} \exp(x \cdot x^-_n / \theta)},$}
\end{equation}
where $N$ is the number of negative samples and $\theta$ is a temperature hyper-parameter. 

However, the assumption of uniform degradation across the image does not always hold due to lighting, local motion, defocus, and other factors. For example, it is possible to have a moving face with a static background in an image, which means that only some patches suffer from motion blur. This oversimplified assumption often leads to suboptimal and inconsistent results for degradation learning.

\subsubsection{Our Solution}
\label{sec:oursol}
To bridge this gap, we propose Dual-Set Degradation Representation Learning (DSL), which considers entire face images. To make this challenging setting compatible with contrastive learning approaches, we carefully construct two sets of images, $\mathcal{S}$ and $\mathcal{R}$, each serving a unique purpose in the degradation learning process, as shown in \figref{fig:dce}.

\textit{Set $\mathcal{S}$} consists of a collection of images derived from a single high-quality face image, with each image undergoing different types of $\mathcal{S}$ynthetic degradation
including but not limited to blurring, noise, resizing, JPEG compression, and extreme lighting conditions. This set acts as a controlled environment, enabling in-depth exploration of a wide \emph{variety of degradations} against \emph{constant} content.

In contrast, \textit{Set $\mathcal{R}$} encompasses a compilation of images from GFIQA datasets, each having \emph{different} content under $\mathcal{R}$\emph{eal-world} degradation. This set reflects the unpredictability and diversity of realistic degradations, which are hard to model by synthetic data.

Formally, let \( \mathcal{S} = \{ s_1, \dots, s_m \} \) and \( \mathcal{R} = \{ r_1, \dots, r_n \} \), where \( m \) and \( n \) represent the number of images in \(\mathcal{S}\) and \(\mathcal{R}\), respectively. Each image from the two sets is mapped to its degradation representation by a function $\psi$ defined by the degradation extraction module with weights $z$:
\begin{equation}
\scalebox{0.95}{$
\psi_z^\rightarrow(\mathcal{S}) = \{ \psi(s_1;z), \dots, \psi(s_m;z) \}$}
\end{equation}
\begin{equation}
\scalebox{0.95}{$
\psi_z^\rightarrow(\mathcal{R}) = \{ \psi(r_1;z), \dots, \psi(r_n;z) \}$}
\end{equation}

Inspired by~\cite{rocco2018neighbourhood,snell2017prototypical}, we introduce a mechanism termed \emph{soft proximity mapping}: For a given image $s_i$ from $\mathcal{S}$, we map its representation $\psi(s_i)$ to a linear combination of representaions in $\psi_z^\rightarrow(\mathcal{R})$ as follows:
\begin{equation}
\scalebox{1.0}{$
\hat{\psi}(s_i) = \sum_{j=1}^n \text{sim}(\psi(s_i), \psi(r_j)) \cdot \psi(r_j)$}
\end{equation}
where $\hat{\psi}(s_i)$ denotes soft proximity mapping of $\psi(s_i)$. \( \text{sim}(\cdot, \cdot) \) denotes the similarity between two representations. We use $\mathcal{L}2$ distance as our similarity metric in our implementation. $z$ is omitted for brevity.

This construction allows us to define positive and negative pairs for contrastive learning. Intuitively, a degradation representation $\psi(s_i)$ should be attracted to its own soft proximity mapping $\hat{\psi}(s_i)$, while any other representations $\psi(s_j)$ where $j\neq i$ should be repelled from this soft proximity mapping because $s_i$ and $s_j$ have different degradations by the dedicated construction of \textit{Set $\mathcal{S}$}. Then, we adopt the contrastive loss~\cite{oord2018representation}:\begin{equation}
\scalebox{0.94}{$
\mathcal{L}_{Con}(\mathcal{S}, \mathcal{R})=-\frac{1}{m} \sum_{i=1}^m \log \frac{\textnormal{exp}(\text{sim}(\psi(s_i),\hat{\psi}(s_i))/\theta)}{\sum_{j\neq i}^n \textnormal{exp}(\text{sim}(\psi(s_j),\hat{\psi}(s_i))/\theta)}
$}
\end{equation}

This loss function leverages the nature that within \( \mathcal{S} \), images share the same \emph{content} but differ in \emph{degradations}, contrasting with \( \mathcal{R} \), which varies in both aspects.
By drawing the extracted degradation representation closer to its corresponding soft proximity mapping and distancing it from other soft proximity mappings, the degradation extraction module is trained to learn a global degradation representation that is independent of the image content.

Furthermore, the self-supervised \textbf{dual-set} contrastive learning strategy is essential for understanding various degradations, particularly in real-world scenarios. This approach is vital as it involves accurately extracting degradation representations from real-world images to approximate those in the synthetic set \( \mathcal{S} \). It is feasible to employ contrastive learning solely on the synthetic set \( \mathcal{S} \) to capture degradation patterns: Positive pairs consist of images with the same degradation, and negative pairs otherwise. However, this naive approach does not generalize well to real-world images. In contrast, our dual-set design can bring together the benefits of both the synthetic set with controllable degradations and the real-world set with realistic degradations, achieving better generalization.

Notice that the roles of \( \mathcal{S} \) and \( \mathcal{R} \) are symmetric:

Just as we utilize representations from \( \mathcal{S} \) to seek corresponding features within \( \mathcal{R} \), empirically, we found the reverse is also viable and informative. Thus, we define our Degradation Extraction Loss $\mathcal{L}_{DE}$ as a bidirectional loss:
\begin{equation}
\scalebox{1.0}{$
\mathcal{L}_{DE}=\mathcal{L}_{Con}(\mathcal{S}, \mathcal{R})+\mathcal{L}_{Con}(\mathcal{R}, \mathcal{S})$}
\label{eq:lde}
\end{equation}
This bidirectional loss reinforces the mutual learning and alignment between the synthetic and real-world sets, ensuring a comprehensive understanding and representation of realistic degradations. Moreover, it is worth mentioning that the high-quality image in set \( \mathcal{S} \) is resampled for every iteration, where this image undergoes random synthetic degradations of varying intensities. Concurrently, images in set \( \mathcal{R} \) are also resampled randomly in each iteration.
 
In summary, DSL gets rid of the uniformity assumption of degradation in patches across the entire image for degradation learning. Instead, it relies on the soft proximity mapping between two constructed sets of images to calculate the contrastive loss, which allows for more precise degradation representation (this mechanism is kind of similar in spirit to~\cite{murphy2022learning}). Furthermore, since the entire image is considered, DSL can capture a holistic view of the degradation unique to each image, further boosting the performance.

\subsection{Landmark-guided GFIQA}
Face images are uniquely challenging in image processing. This is because human eyes are especially sensitive to facial artifacts, raising the importance of nuanced quality assessment~\cite{wang2022survey}. Thus, it is important to design an approach that does not treat each pixel equally; it should acknowledge the perceptual significance of salient facial features. Furthermore, as stated in~\cref{sec:model_overview}, considering that our network crops the face into various patches to compute the average MOS score, it is crucial to provide landmark information to give the spatial context on which part of the face each patch covers, ensuring a holistic and perceptually consistent evaluation.

As shown in \figref{fig:overview}, our approach begins with utilizing an existing landmark detection algorithm (\textit{i.e.,} 3DMM model~\cite{egger20203d}) to identify key facial landmarks. Inspired by Neural Radiance Fields (NeRF)~\cite{mildenhall2021nerf}, we apply positional encoding to these unique landmark identifiers. 
By applying a series of sinusoidal functions to the raw identifiers, positional encoding enhances the representational capacity of the network, allowing the network to capture and learn more intricate relationships and patterns associated with each landmark identifier.

The encoded information is subsequently concatenated with the features processed by the Transformer Decoder, feeding into the regional confidence branch. The human visual system is particularly sensitive to high-frequency details, which are often associated with facial landmarks such as the eyes, nose, and mouth. Providing this landmark-based information to the confidence head can help generate a more precise confidence map, emphasizing regions that humans naturally prioritize in their perception.

In our approach, we deliberately avoid relying on encoding landmark coordinates $(x,y)$ in an image as positions, as it can introduce ambiguity during learning, especially when faces are unaligned, or images are cropped into patches. In such scenarios, specific coordinates may inconsistently correspond to different facial features on different training samples, therefore muddling the learning process.
To avoid this, our network employs a fixed encoding scheme for each facial landmark, assigning a unique identifier to every critical feature regardless of its position in the image. This methodology proves particularly advantageous for our ViT, which takes fixed-size crops from the input image, potentially capturing only portions of the face.

Given the diverse range of degradations encountered in GFIQA, off-the-shelf landmark detectors often fail on images with challenging degradations. We observed that fine-tuning existing landmark detectors like \cite{jin2021pixel,kartynnik2019real,deng2020retinaface} on degraded images leads to more accurate landmark detection.

In summary, by adopting landmark-guided cues, our method maintains a consistent awareness of crucial facial features within each crop, which effectively encourages the model to focus on salient facial features when aggregating the regional quality scores.

\subsection{Loss Functions}
\noindent\smallskip
\textbf{Degradation Encoder.}
The degradation encoder is trained separately by optimizing \cref{eq:lde}. Once trained, it remains fixed when training the core GFIQA network. 
\noindent\smallskip\\
\textbf{GFIQA Network.} To measure the discrepancy between the predicted MOS and the ground truth, we employ the Charbonnier loss~\cite{lai2017deep} ($\mathcal{L}_{char}$), which is defined as:

\begin{equation}
\mathcal{L}_{char}(p, \hat{p}) = \sqrt{(p - \hat{p})^2 + \epsilon^2}
\end{equation}
\noindent where \( \hat{p} \) is the predicted MOS, \( p \) is the ground truth MOS, and \( \epsilon \) is a small constant to ensure differentiability. 

Unlike existing GIQA~\cite{yang2022maniqa,ke2021musiq,saha2023re} or GFIQA~\cite{su2023going} models that typically rely on $\mathcal{L}2$ losses, we opt for the Charbonnier loss as it is less sensitive to outliers, which in the context of GFIQA can arise from rare face quality degradations, dataset annotation discrepancies, or occasional extreme scores predicted by the model during training. By improving the robustness against outliers, our model is more aligned with human perceptual judgments.

\section{Comprehensive Generic Face IQA Dataset}

Existing GFIQA models are evaluated on datasets such as PIQ23 \cite{Chahine_2023_CVPR} and GFIQA-20k~\cite{su2023going}. While PIQ23 contains a variety of in-the-wild images with uncropped faces, its constrained dataset size limits its efficacy for training robust models. Moreover, both datasets exhibit biases in \emph{gender} and \emph{skin tone} representation. This disproportion can introduce biases during model training, decreasing the performance and reliability of models in face image quality assessment tasks. Prior research~\cite{buolamwini2018gender,selbst2019fairness,klare2012face,cook2019demographic} has shown that this imbalance in data distribution has a significant negative impact on model performance in various face-related applications. 

To tackle these challenges, we introduce a new dataset named Comprehensive Generic Face Image Quality Assessment (CGFIQA-40k), which includes approximately 40K images, each with a resolution of 512x512. Each image is annotated by 20 labelers, and each labeler spends about 30 seconds to give a score. From an initial pool of 40,000 images, we filtered out a small number of images with unusable content or incomplete labels, resulting in a total of 39,312 valid images. This dataset is specifically curated to include an extensive collection of face images with diverse distribution on skin tone, gender, and facial occlusions such as masks and accessories.

A comparative overview of our dataset with existing GFIQA datasets is provided in~\tabref{tab:dataset}. We hope this dataset will offer a more comprehensive benchmark for GFIQA, pushing the generalization and robustness of state-of-the-art methods.

\begin{table}
\centering
\scalebox{0.7}{
\begin{tabular}{ccccccc} 
\toprule
\multirow{2}{*}{Dataset} & \multirow{2}{*}{Size} & \multicolumn{3}{c}{Skin Tone ($\%$)} & \multicolumn{2}{c}{Gender ($\%$)}  \\
\cmidrule(lr){3-5} \cmidrule(lr){6-7}
                        &                       & Light & Medium & Dark & Male & Female \\ 
\hline\hline
PIQ23              & 5116                  & 74.5    & 9.0 & 16.5  & 94.0 & 6.0 \\
GFIQA-20k               & 20000            & 81.6    & 6.7 & 11.7  & 64.2 & 35.8 \\
CGFIQA-40k                  & \textbf{39312}            & \textbf{53.83}    & \textbf{24.91} & \textbf{21.26} & \textbf{51.50} & \textbf{48.50} \\
\bottomrule
\end{tabular}
}
\caption{\textbf{Comparison of datasets in terms of size, skin tone, and gender distribution.} Skin tones are categorized as Light (Fitzpatrick scale~\cite{fitzpatrick1975soleil} I-II), Medium (Fitzpatrick scale III-IV), and Dark (Fitzpatrick scale V-VI).}
\label{tab:dataset}
\end{table}

\section{Experimental Results}
\subsection{Experiment Settings}
\label{sec:implementation}
Our experiments utilize three datasets: GFIQA-20k~\cite{su2023going}, PIQ23~\cite{Chahine_2023_CVPR}, and our newly introduced CGFIQA-40k. The GFIQA-20k dataset consists of 20,000 images, divided into 14,000 for training, 2,000 for validation, and 4,000 for testing. PIQ23 contains 5,116 images, with 3,581 for training, 512 for validation, and 1,023 for testing. Our CGFIQA-40k dataset includes a more extensive collection of 39312 images, with a division of 27518 for training, 3931 for validation, and 7863 for testing.

The evaluation metrics employed are the Pearson Linear Correlation Coefficient (PLCC) and Spearman's Rank Order Correlation Coefficient (SRCC).

\begin{table}[t!]
\small
\begin{center}
\scalebox{0.7}{
\begin{tabular}{ccccccccc}
\toprule
\multirow{2}{*}{\textbf{Method}} & \multicolumn{2}{c}{\textbf{GFIQA-20k}} & \multicolumn{2}{c}{\textbf{PIQ23}} & \multicolumn{2}{c}{\textbf{CGFIQA-40k}} \\
\cline{2-7}
& PLCC & SRCC & PLCC & SRCC & PLCC & SRCC \\
\hline\hline
\rowcolor{lightgray}
ArcFace~\cite{deng2019arcface} & 0.9508 & 0.9510 & 0.5913 & 0.6011 & 0.9722 & 0.9723 \\
\rowcolor{lightgray}
MegaFace~\cite{meng2021magface} & 0.9523 & 0.9531 & 0.5941 & 0.5355 & 0.9731 & 0.9733 \\
\rowcolor{lightgray}
CR-FIQA~\cite{boutros2023cr} & 0.9593 & 0.9598 & 0.6013 & 0.6021 & 0.9734 & 0.9736 \\
\rowcolor{pink}
Koncept512~\cite{hosu2020koniq} & 0.9518 & 0.9523 & 0.6013 & 0.6007 & 0.9713 & 0.9721 \\
\rowcolor{pink}
MUSIQ~\cite{ke2021musiq} & 0.9503 & 0.9518 & 0.7141 & 0.7101 & 0.9750 & 0.9735 \\
\rowcolor{pink}
ReIQA~\cite{saha2023re} & 0.9437 & 0.9446 & 0.5988 & 0.5961 & 0.9800 & 0.9802 \\
\rowcolor{pink}
CONTRIQUE~\cite{madhusudana2022image} & 0.9458 & 0.9466 & 0.5892 & 0.5930 & 0.9788 & 0.9799 \\
\rowcolor{pink}
UNIQUE~\cite{zhang2021uncertainty} & 0.9413 & 0.9528 & 0.5822 & 0.5710 & 0.9771 & 0.9641 \\
\rowcolor{pink}
MANIQA~\cite{yang2022maniqa} & 0.9614 & 0.9604 & \underline{0.7202} & \underline{0.7180} & 0.9805 & 0.9809 \\
\rowcolor{pink}
TReS~\cite{golestaneh2022no} & 0.9512 & 0.9552 & 0.5767 & 0.5760 & 0.9816 & 0.9817 \\
\rowcolor{pink}
HyperIQA~\cite{su2020blindly} & 0.9664 & 0.9674 & 0.7152 & 0.7203 & 0.9722 & 0.9733 \\
\rowcolor{pink}
LIQE~\cite{zhang2023liqe} & 0.9341 & 0.9571 & 0.4627 & 0.4522 & 0.9786 & 0.9623 \\
\rowcolor{pink}
MetaIQA~\cite{zhu2020metaiqa} & 0.9542 & 0.9532 & 0.6008 & 0.6016 & 0.9474 & 0.9463 \\
\rowcolor{pink}
TRIQ~\cite{you2021transformer} & 0.9631 & 0.9630 & 0.6118 & 0.6023 & 0.9813 & \underline{0.9821} \\
\rowcolor{pink}
VCRNet~\cite{pan2022vcrnet} & 0.9672 & 0.9679 & 0.6826 & 0.6774 & 0.9819 & 0.9821 \\
\rowcolor{pink}
GraphIQA~\cite{sun2022graphiqa} & 0.9434 & 0.9436 & 0.4676 & 0.4709 & 0.9578 & 0.9552 \\
\rowcolor{LightCyan}
IFQA~\cite{Jo_2023_WACV} & 0.9601 & 0.9603 & 0.2907 & 0.3081 & 0.9791 & 0.9803 \\
\rowcolor{LightCyan}
StyleGAN-IQA~\cite{su2023going} & \underline{0.9673} & \underline{0.9684} & 0.7013 & 0.7131 & \underline{0.9822} & \underline{0.9821} \\
\textbf{DSL-FIQA} & \textbf{0.9745} & \textbf{0.9740} & \textbf{0.7370} & \textbf{0.7333} & \textbf{0.9873} & \textbf{0.9880} \\
\bottomrule
\end{tabular}}
\end{center}
\caption{\textbf{Comparison of various image quality assessment methods across three GFIQA datasets.} \textbf{Boldface} indicates the best results and \underline{underline} the second-best. Our approach demonstrates superior and robust performance across all three datasets. Rows in gray denote BFIQA methods, those in pink for generic IQA methods, and in light cyan for GFIQA methods.}
\label{tab:performance}
\end{table}

\begin{table}[t!]
\centering
\scalebox{0.6}{
\begin{tabular}{cccccccc|ccc} 
\toprule
\multicolumn{8}{c}{\textbf{Module}} & \multicolumn{2}{c}{\textbf{Metric}}  \\ 
\hline
Patch&DSL-cat & DSL-S & DSL-R & DSL &Landmark & PE & $\mathcal{L}_{Char}$ & PLCC & SRCC \\
\hline\hline
- &- & - &- &- &- & - & -  &  0.9682 & 0.9679   \\
\checkmark &- &- & - &- &- & - & -  &  0.9695 & 0.9687   \\
- &\checkmark &- &- & - &- & - & -   &  0.9703 & 0.9701   \\
- &- & \checkmark &- &- & - & - & -  & 0.9713 & 0.9711 \\
- &- & - & \checkmark &- & - & - & -  & 0.9709 & 0.9707 \\
- &- & - & - & \checkmark &- & - & -  & 0.9721 & 0.9719 \\
- &- & - & - &\checkmark & \checkmark &- &-  & 0.9731 & 0.9728 \\
- &- & - & - &\checkmark & \checkmark & \checkmark  &-  & 0.9735 & 0.9731 \\
- &- & -&- &\checkmark & \checkmark & \checkmark& \checkmark  & \textbf{0.9745} & \textbf{0.9740}  \\
\bottomrule
\end{tabular}}
\\
\caption{\textbf{Ablation study on proposed techniques.} We demonstrate that the proposed modules can effectively improve GFIQA's performance. ``PE'' denotes the positional encoding operation. ``DSL-S'' and ``DSL-R'' represent Dual-Set Degradation Representation Learning. In this framework, the degradation encoder is trained using a \emph{one-sided} loss approach, specifically $\mathcal{L}_{Con}(\mathcal{S}, \mathcal{R})$ for ``DSL-S'' and $\mathcal{L}_{Con}(\mathcal{R}, \mathcal{S})$ for ``DSL-R''. ``DSL-cat'' refers to directly concatenating the degradation representations with image features instead of cross-attention, while ``DSL'' indicates the adoption of a \emph{dual-sided} loss function as defined in \cref{eq:lde}.} 
\label{tab:ablation}
\end{table}

\begin{figure}[t!]
  \centering
  \begin{subfigure}[b]{0.52\columnwidth} 
    \includegraphics[width=\textwidth]{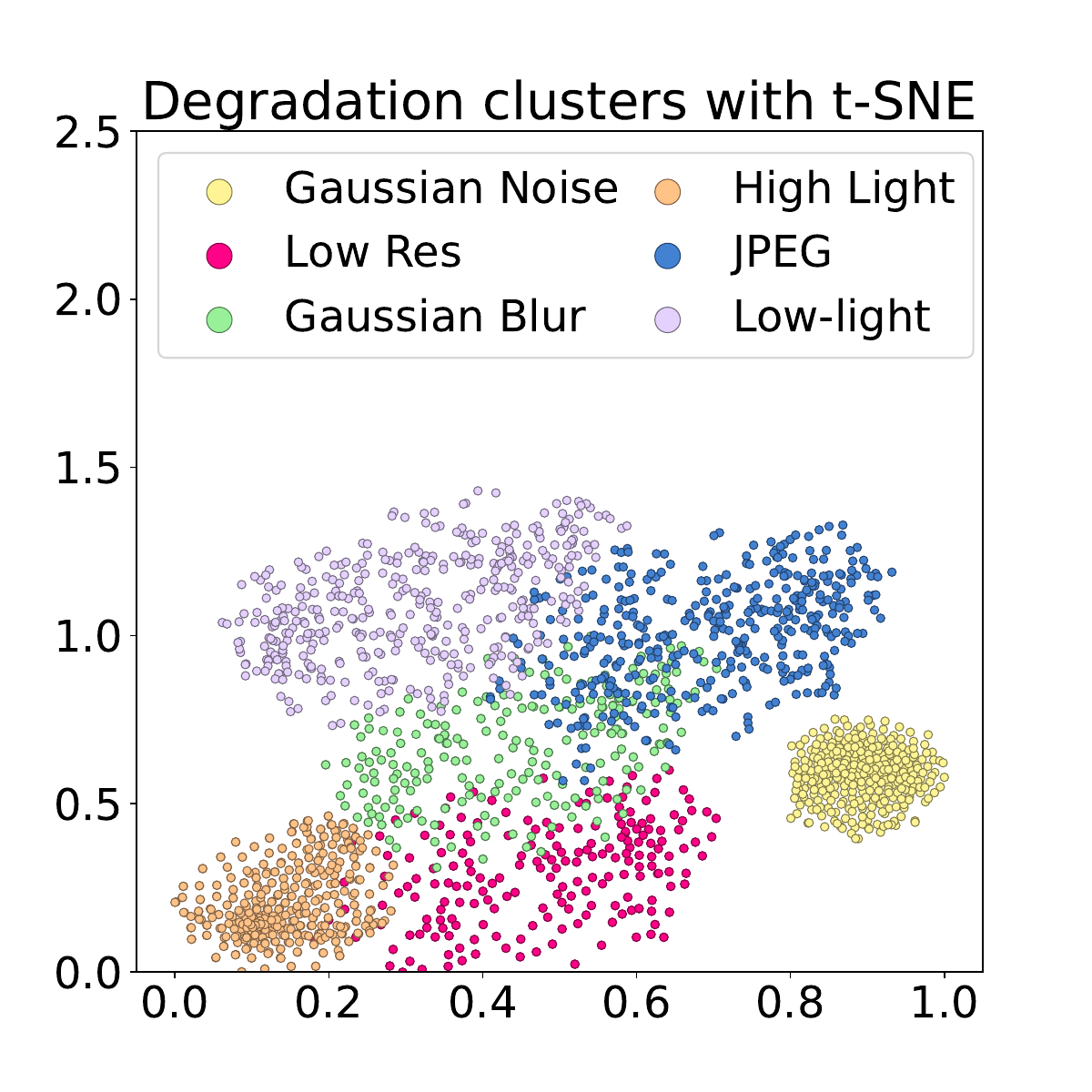} 
    \caption{Patch-based}
    \label{fig:subim1}
  \end{subfigure}
  \hspace{-0.6cm} 
  \begin{subfigure}[b]{0.52\columnwidth} 
    \includegraphics[width=\textwidth]{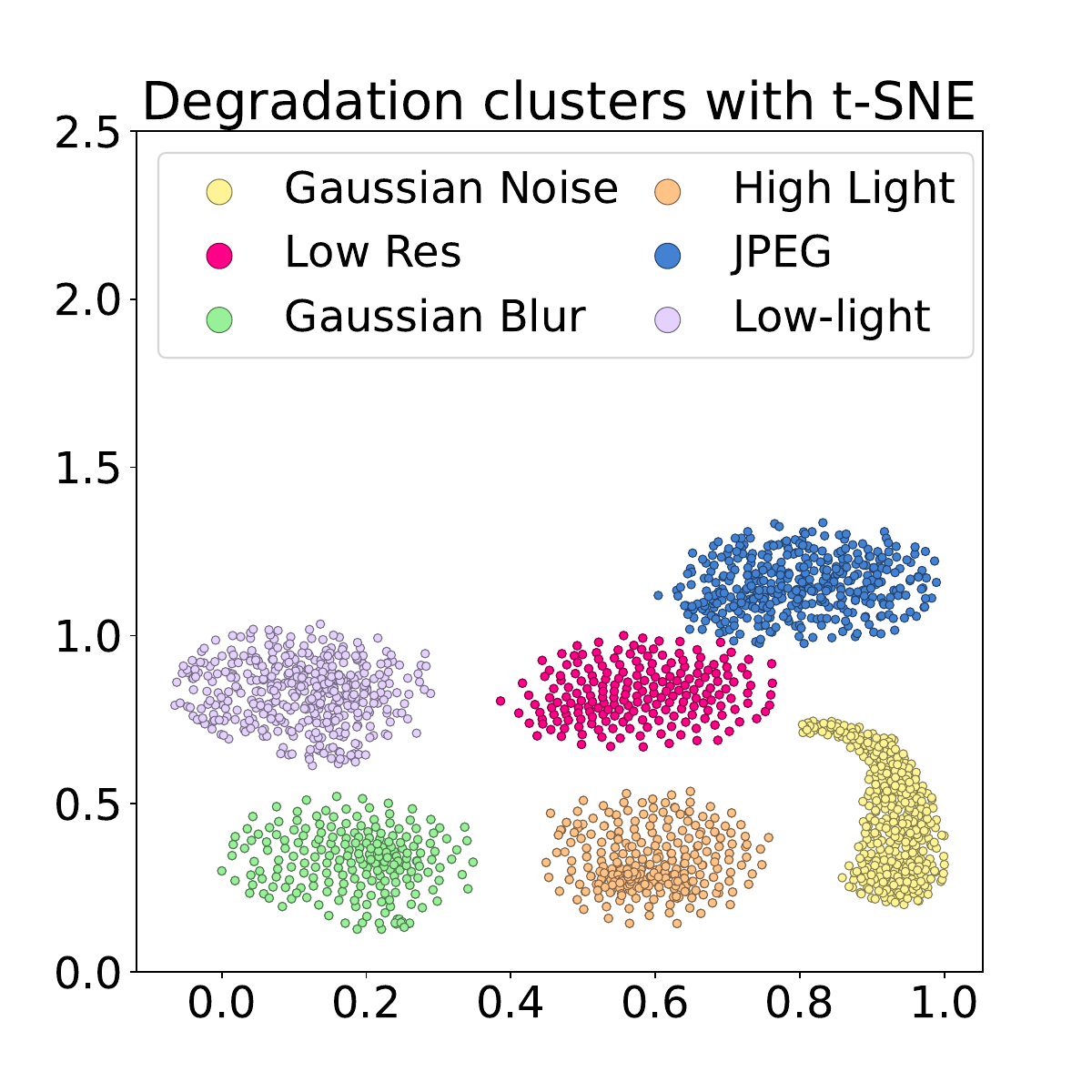} 
    \caption{DSL}
    \label{fig:subim2}
  \end{subfigure}
  \caption{\textbf{t-SNE visualization of degradation representation extracted using patch-based and DSL-based methods.} Unlike the patch-based method, DSL results in well-demarcated clusters for various types of degradation, thereby proving the effectiveness of the learned representations.}

  \label{fig:ablation_tsne}
\end{figure}

\begin{table}[t!]
\centering
\scalebox{0.8}{
\begin{tabular}{ccc} 
\toprule
\textbf{Strategy }      & Patch-based &DSL \\ 
\hline\hline
mAP &      48.2  & \textbf{69.3}           \\
\bottomrule
\end{tabular}}
\\
\caption{\textbf{Comparative retrieval accuracy using DSL and patch-based methods under real-world degradations, quantified by mAP scores.}}
\label{tab:ablation-DE}
\end{table}

\subsection{Performance Evaluation}
We conducted an extensive evaluation of our method via experiments, comparing it with representative models across the datasets GFIQA-20k, PIQ23, and CGFIQA-40k. To maintain a fair comparison, all models were trained and validated under the identical conditions specified in \cref{sec:implementation}.

We compare with a wide range of generic IQA models, including Koncept512~\cite{hosu2020koniq}, MUSIQ~\cite{ke2021musiq}, ReIQA~\cite{saha2023re}, CONTRIQUE~\cite{madhusudana2022image}, UNIQUE~\cite{zhang2021uncertainty}, MANIQA~\cite{yang2022maniqa}, TReS~\cite{golestaneh2022no}, HyperIQA~\cite{su2020blindly}, LIQE~\cite{zhang2023liqe}, MetaIQA~\cite{zhu2020metaiqa}, TRIQ~\cite{you2021transformer}, VCRNet~\cite{pan2022vcrnet}, and GraphIQA~\cite{sun2022graphiqa}. We also compare with recent GFIQA methods, including StyleGAN-IQA~\cite{su2023going} and IFQA~\cite{Jo_2023_WACV}. For completeness, we include three representative BFIQA methods, ArcFace~\cite{deng2019arcface}, MegaFace~\cite{meng2021magface}, and CR-FIQA~\cite{boutros2023cr}.

\tabref{tab:performance} clearly shows the robust performance of our method, which outperforms existing models across all metrics on GFIQA-20k, PIQ23, and CGFIQA-40k. The consistent results across diverse datasets validate our approach's strength and adaptability, establishing it as a robust generic face image quality assessment solution.

\begin{figure}[t!]
  \centering
  \begin{subfigure}[b]{0.286\columnwidth} 
    \includegraphics[width=\textwidth, height=6.8cm]{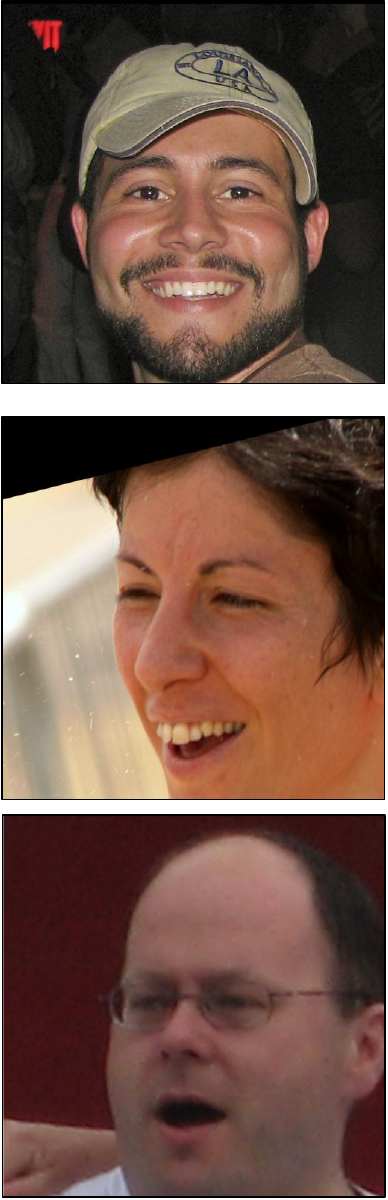} 
    \caption{Input}
    \label{fig:subim1}
  \end{subfigure}
  \hspace{0.001cm} 
  \begin{subfigure}[b]{0.337\columnwidth} 
    \includegraphics[width=\textwidth, height=6.8cm]{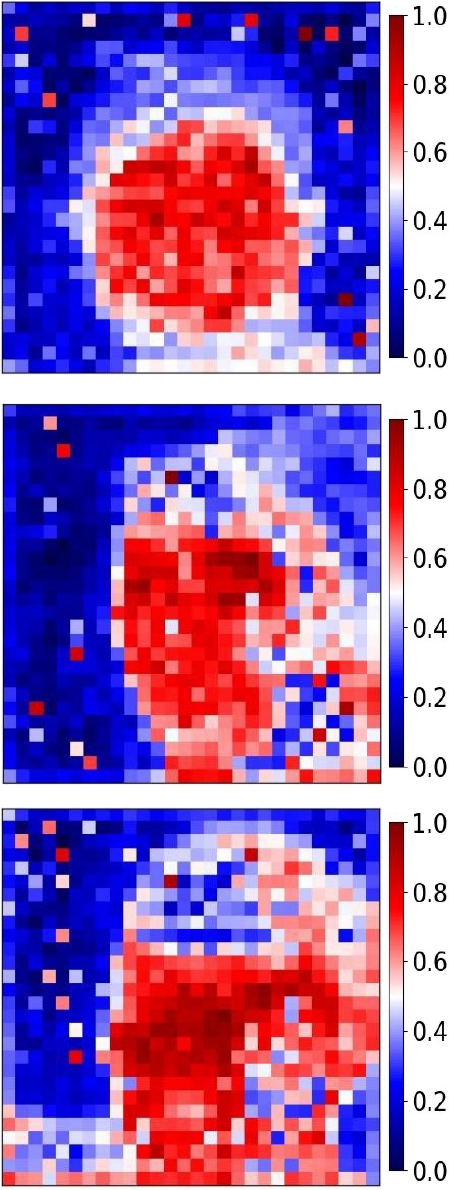} 
    \caption{w/o landmark}
    \label{fig:subim2}
  \end{subfigure}
  \hspace{0.001cm} 
  \begin{subfigure}[b]{0.337\columnwidth} 
    \includegraphics[width=\textwidth, height=6.8cm]{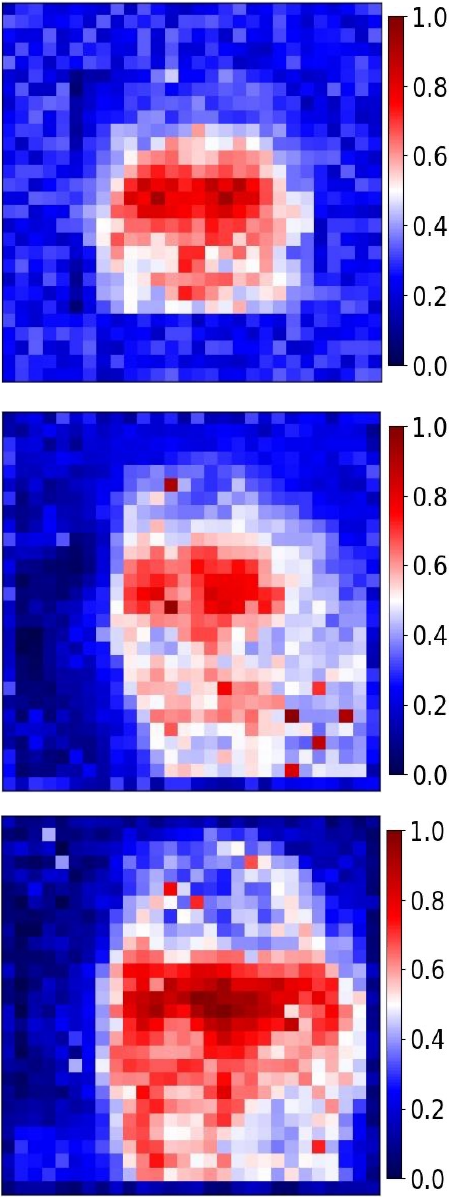}  
    \caption{w landmark}
    \label{fig:subim3}
  \end{subfigure}
\caption{\textbf{Comparison of using landmark mechanism to guide the GFIQA network.} We present the regional confidence maps and the corresponding input. With landmark guidance, the confidence maps focus more on key facial landmarks, providing a more discriminative assessment. In contrast, without landmark guidance, the confidence maps tend to cover the entire face, often lacking specificity and even assigning higher confidence to irrelevant areas (e.g., background).}

  \label{fig:ablation_landmark}
\end{figure}

\subsection{Ablation Study}
To validate the effectiveness of individual components of our approach, we conducted an ablation study based on the GFIQA-20k dataset.
\noindent \smallskip\\
\textbf{Effectiveness of Degradation Extraction.} 
The innovation of our method lies in the Dual-set Degradation Representation Learning (DSL), enhancing GFIQA precision by capturing global degradation in degradation learning. To verify its impact, we show the comparison in \tabref{tab:ablation}. Including DSL improved the performance of the GFIQA (Row 1 and Row 6), highlighting its critical role in boosting GFIQA accuracy. We also substituted the DSL learning technique for the existing patch-based degradation learning in \tabref{tab:ablation} (Row 2 and Row 6), and the results indicate that the proposed DSL can improve GFIQA performance more than the patch-based strategy (\cref{sec:patch_based}). We also validated the significance of cross-attention for integrating degradation information (Row 3 and Row 6). The results indicated that employing cross-attention for this integration yielded superior outcomes. 
\noindent \smallskip\\
\textbf{Comparison between DSL and patch-based methods.} To further evaluate the advantage of our DSL over patch-based degradation extraction methods, we conducted two additional experiments. Both methods were trained on the same dataset for fair and direct comparison. The testing data used was completely independent of the training set, guaranteeing the validity of our assessment.

The first experiment involved synthetic data. We randomly selected 1,000 face images from the FFHQ dataset~\cite{karras2019style}, subjecting each to six types of synthetic degradations. We then employed DSL and patch-based feature extraction, subsequently visualizing these features using t-SNE. The results, illustrated in~\figref{fig:ablation_tsne}, demonstrated that DSL could effectively separate the images based on their specific degradations, while the patch-based method showed considerable overlap.

Furthermore, we extended our exploration to real-world conditions, using images from the GFIQA-20k dataset. This second experiment was designed to verify if the distinct degradation representations learned by DSL could enhance image degradation retrieval accuracy under real-world degradations. To this end, we synthesized six types of degradations on 100 images from the FFHQ dataset. These synthetically degraded images were used as queries to probe the GFIQA-20k test set, selecting the top 5 images with the smallest distance. We then verify whether the five images fall under the same degradation category by human inspection. We quantified our method's precision by calculating the mean average precision (mAP) for these retrieval tasks, as shown in \tabref{tab:ablation-DE}. The results confirmed DSL's enhanced accuracy in identifying images with similar degradation attributes.

In conclusion, these experiments emphasized the effectiveness of DSL in crafting distinct degradation representations and its practical superiority in real-world scenarios, bolstering its value in improving GFIQA outcomes.
\noindent \smallskip\\
\textbf{Effectiveness of Landmark-guided GFIQA.} Integrating facial landmarks into GFIQA significantly improves quality assessment accuracy, addressing the complexity of facial features often ignored in traditional methods, which is validated in \tabref{tab:ablation} (Row 6 and 7). To understand how the landmark guidance works, \figref{fig:ablation_landmark} visualizes the regional confidences predicted with and without landmarks guidance. When landmarks are not used, the model indiscriminately overemphasizes the entire face and even background areas. In contrast, the model with landmark guidance focuses on crucial facial regions, which are more aligned with human perception. In addition, \tabref{tab:ablation} (Row 7 and 8) substantiate the benefit of applying positional encoding to the landmark identifiers, showing that positional encoding can indeed enhance the model capacity to capture more complex relationships inherent in facial features, thereby improving the overall prediction accuracy. 
\noindent \smallskip\\
\textbf{Effectiveness of Charbonnier Loss.} 
The introduction of Charbonnier loss improves the accuracy of GFIQA as shown in \tabref{tab:ablation} (Row 8 and 9).

\section{Conclusion}
In this paper, we tackle the inherent complexities in GFIQA with a transformer-based method. Our Dual-Set Degradation Representation Learning improves degradation extraction, and the additional guidance from facial landmarks further improves the assessment accuracy. Furthermore, we curate the CGFIQA-40k Dataset, rectifying imbalances in skin tones and gender ratios prevalent in previous datasets. Extensive experimental results show that the proposed method performs favorably against state-of-the-art methods across several GFIQA datasets.

\clearpage

\appendix

\renewcommand{\theequation}{S.\arabic{equation}}
\renewcommand\thefigure{S.\arabic{figure}}
\renewcommand\thetable{S.\arabic{table}}
\setcounter{equation}{0}
\setcounter{figure}{0}
\setcounter{table}{0}

{
\twocolumn
\begin{center}
\textbf{\large Supplemental Materials}
\end{center}
\section{Social Impact and Ethical Considerations}
This paper contributes to the advancement of Face IQA technology, which has widespread applications in digital media and social networking platforms. By ensuring a balanced representation of gender and skin tones in the dataset, this paper addresses critical issues of fairness and bias in AI, promoting more equitable facial analysis technologies.

However, if the facial image quality assessment (IQA) method fails, it could lead to the selection of incorrect facial quality images for training, subsequently affecting the accuracy of facial-related algorithms trained on these images. This situation might result in biases or errors in facial recognition, emotion analysis, or other applications based on facial images.

To address this issue, an effective approach is to double-check the images filtered through the Face IQA method to ensure their quality meets the expected standards. This can be achieved through manual review or by employing additional verification mechanisms. Such a double-checking mechanism helps reduce the risk of erroneously selected images, ensuring the quality of training data, thereby enhancing the reliability and effectiveness of algorithms trained on these data.
\section{Comprehensive Generic Face IQA Dataset}
\subsection{Data Collection}
To create a diverse and comprehensive dataset, we initially collected face images from the CelebA dataset. We utilized skin tone~\cite{bevan2022detecting} and gender~\cite{genderdetection} detectors to analyze these images, ensuring a balanced representation of both gender and skin tones. This careful sampling approach was complemented by the addition of 1,028 images from Flickr, specifically chosen to enhance the diversity in terms of skin tones and occlusion. The combined dataset consists of about 40,000 images. Each image was aligned using Dlib's face landmark detection \cite{king2009dlib,kazemi2014one,sagonas2016300} according to FFHQ dataset \cite{karras2019style} and subsequently rescaled to a uniform resolution of $512\times512$ pixels, ensuring consistency across the dataset.

\begin{figure*}[t!]
\centering \includegraphics[width=0.92\textwidth,page=1]{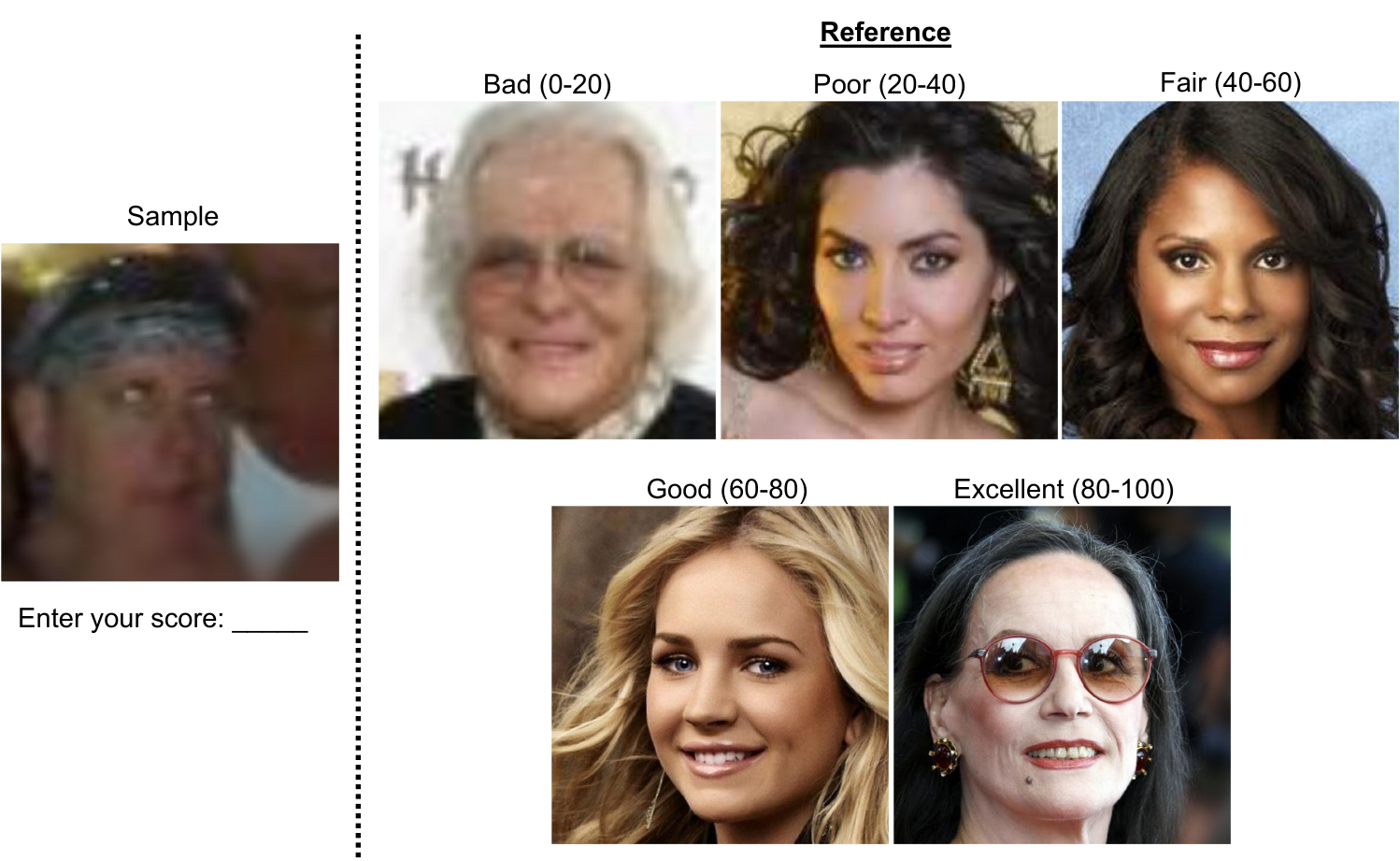}{}
\makeatother 
\caption{\textbf{User Interface of the Subjective Generic Face IQA Study.} Participants assess each image's visual quality by entering the scores in a toolbox.}  
\label{fig:label_ui}
\end{figure*}

\begin{figure}[t!]
\centering \includegraphics[width=0.40\textwidth,page=1]{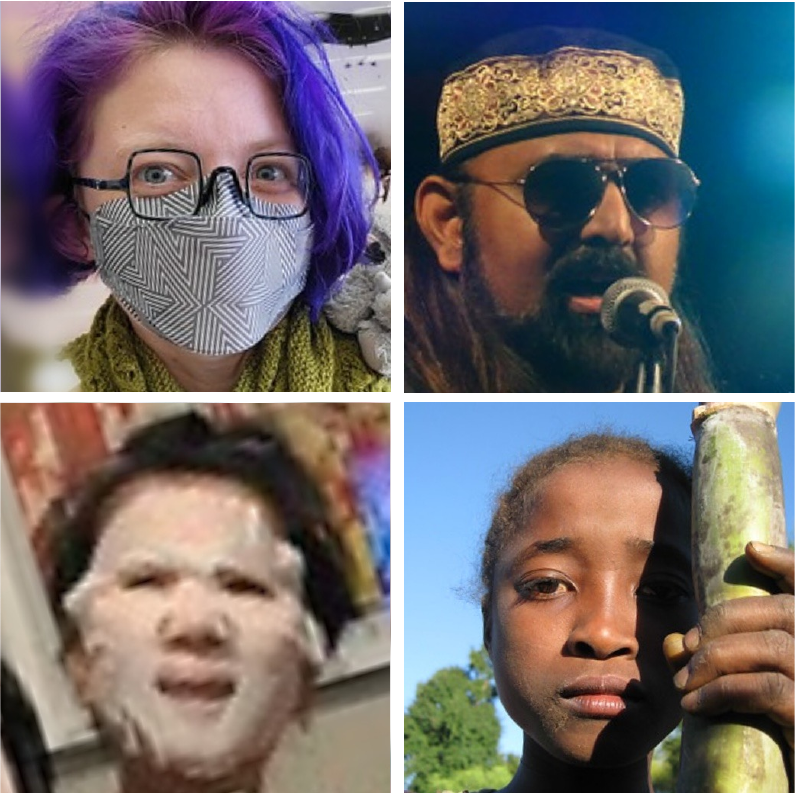}{}
\makeatother 
\caption{\textbf{Examples of occluded images in CGFIQA-40K dataset.}}  
\label{fig:occ_ex}
\end{figure}

\subsection{Annotation Procedure}
To ensure accurate and consistent subjective quality assessment of facial images, we provided annotators with a user-friendly and intuitive interface. This interface was designed to display one facial image at a time, accompanied by an input field for annotators to enter their Mean Opinion Score (MOS) for the image. To assist annotators in making accurate judgments, we included example images for each quality level alongside the interface. These examples served as references, aiding annotators in better discerning and assessing the quality of each image. Additionally, our system supports arbitrary zoom-in and out features for each image, allowing annotators to better assess the details. An illustration of the user interface used in our study is shown in \figref{fig:label_ui}.

For the subjective scoring process, we adopted the standard 5-interval Absolute Category Rating (ACR) scale, comprising levels: Bad, Poor, Fair, Good, and Excellent. This scale was linearly mapped to a range of [0, 1.0], corresponding to the ACR scale as follows: Bad at 0-20\%, Poor at 20-40\%, Fair at 40-60\%, Good at 60-80\%, and Excellent at 80-100\%. 

To elevate the precision and uniformity of the evaluations, we crafted detailed guidelines alongside a collection of definitive gold-standard principles. These encompassed several facets of image analysis, such as the visibility of eyelashes, articulated through specific classification tiers:
\begin{compactitem}
    \item Excellent: No visible artifacts, whether viewed as thumbnails or in original size.
    \item Good: Artifacts discernible solely at original size.
    \item Fair: Minor artifacts noticeable in thumbnail views.
\end{compactitem}
Reference images from the GFIQA-20k dataset were instrumental in guiding the annotators.

Additionally, our guidance provides a structure for using midpoint scores when an image does not clearly fit into a single category. For instance, if an image falls between the ``Poor'' and ``Fair'' categories, a midpoint score of 40 is recommended.

We curated a collection of 35 images carefully selected by experts, where each of the five quality intervals is represented by seven images.
Three images from each level were used as golden samples, which were provided to guide each annotator along with the rating guidelines. Additionally, we conducted a pre-annotation training using the remaining 20 images, with four images from each quality level (It is unknown to the annotators that they were evenly distributed). Annotators were required to achieve an accuracy of at least 80\% in this test to complete their training. To clarify, an annotator's assessment was considered correct if their assigned Mean Opinion Score (MOS) was within a margin of ±15 points from the ground truth MOS score. If this criterion was not met, they were asked to revisit the guidelines and 15 example images and then retake the test until they reached the accuracy threshold. Importantly, annotators were not informed of the correct answers to the test questions throughout the process.

In total, we engaged 20 annotators for this study. On average, each annotator spent approximately 30 seconds assessing the quality of each image. This arrangement ensured both the ratings' efficiency, quality, and consistency. These detailed guidelines and scoring mechanisms ensured that participants could accurately and consistently assess image quality, thereby enhancing our dataset's overall quality and reliability.

\begin{figure*}[t!]
\centering \includegraphics[width=1.0\textwidth,page=1]{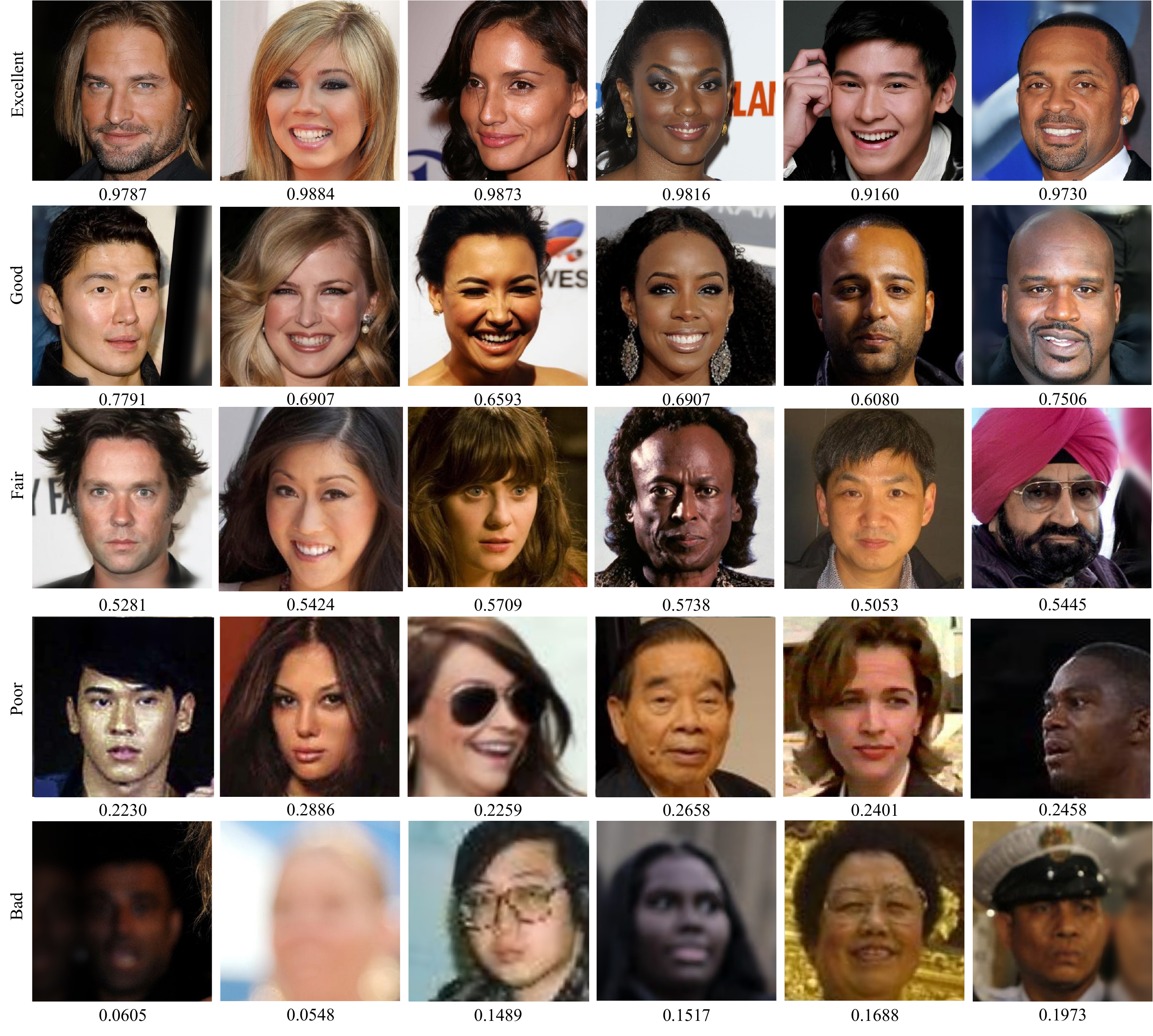}{}
\makeatother 
\caption{\textbf{Sampled face images from the CGFIQA-40k dataset.} These images showcase the diversity of visual quality across five categories: Excellent ((0.8,1]), Good ((0.6,0.8]), Fair ((0.4,0.6]), Poor ((0.2,0.4]), and Bad ([0,0.2]). Each category is represented by six randomly selected images, annotated with their corresponding Mean Opinion Scores (MOS).}  
\label{fig:ex_cgfiqa}
\end{figure*}

\begin{figure}[t!]
\centering \includegraphics[width=0.49\textwidth,page=1]{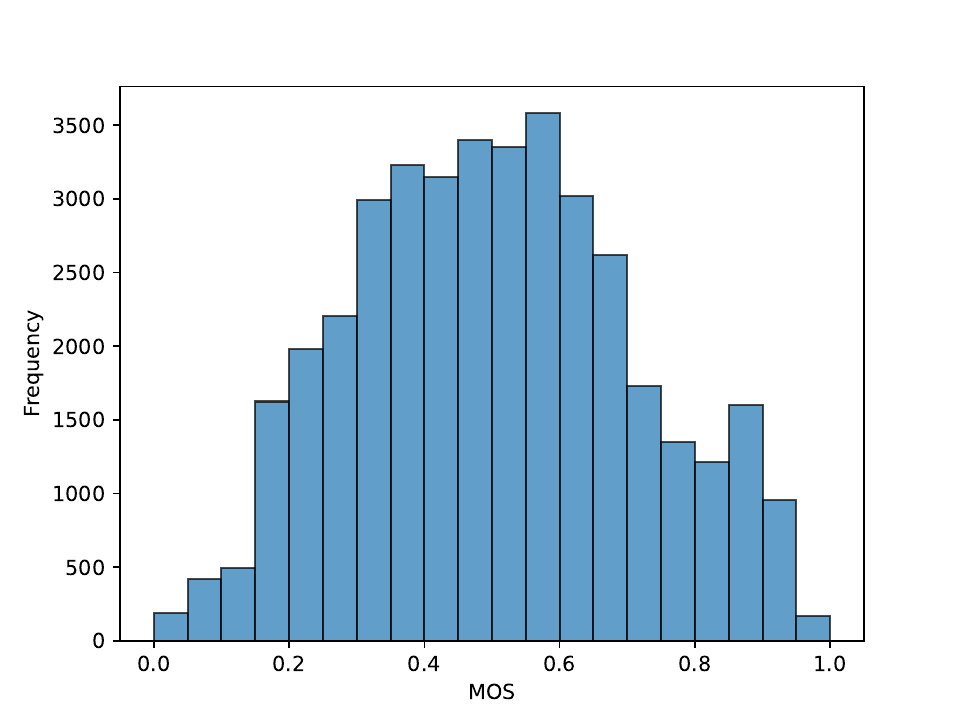}{}
\makeatother 
\caption{\textbf{Distribution of the CGFIQA-40K dataset in terms of MOS scores.}}  
\label{fig:mos_hist}
\end{figure}

\subsection{Dataset Overview}
In this section, we delve into the CGFIQA-40k dataset, which is comprised of 40,000 face images, each meticulously annotated with a Mean Opinion Score (MOS). This dataset represents a comprehensive collection, covering a broad spectrum of image quality with MOS values ranging from 0 to 1.

The CGFIQA-40k dataset is specifically curated to focus on facial images, showcasing various visual qualities, including several images with occlusions. As illustrated in \figref{fig:occ_ex}, these occluded images are integral to the dataset, contributing to its diversity and providing edge cases for robust model training. We have included image samples from different categories - Excellent, Good, Fair, Poor, and Bad to demonstrate the overall diversity. From each category, as shown in \figref{fig:ex_cgfiqa}, six images have been carefully selected to represent the range of qualities within that category. These images and their respective MOS values are displayed in the accompanying figures, illustrating the perceptual quality differences across categories.

Furthermore, we present a histogram of the MOS distribution in \figref{fig:mos_hist} for the entire dataset. This histogram provides a clear overview of the quality distribution of the images, highlighting the frequency and range of different quality levels within the dataset.
\section{Implementation Details}
\subsection{Evaluation Criteria}
In our evaluation, we use two well-established metrics to assess the performance of our model: Spearman's Rank-Order Correlation Coefficient (SRCC) and Pearson's Linear Correlation Coefficient (PLCC).

PLCC measures the linear correlation between actual and predicted quality scores, indicating how closely the predictions align with real values on a linear scale. It is sensitive to numerical differences between scores.

SRCC, in contrast, evaluates the monotonic relationship between two datasets. It focuses on rank order rather than numerical values, offering robustness against outliers and skewed distributions. Both metrics range from -1 to 1, where 1 signifies perfect correlation, -1 indicates perfect inverse correlation, and 0 means no linear correlation. Higher absolute values indicate better performance, with positive values showing consistency with the ground truth.

For the PLCC, given $s_i$ and $\hat{s}_i$ as the actual and predicted quality scores for the $i$-th image, and $\mu_{s_i}$ and $\mu_{\hat{s}_i}$ as their means, with $N$ as the number of test images, it is defined as:
\begin{equation}
    \operatorname{PLCC} =\frac{\sum^{N}_{i=1}(s_i-\mu_{s_i})(\hat{s}_i-\mu_{\hat{s}_i})}{\sqrt{\sum^{N}_{i=1}(s_i-\mu_{s_i})^{2}}\sqrt{\sum^{N}_{i=1}(\hat{s}_i-\mu_{\hat{s}_i})^{2}}},
\end{equation}
For SRCC, where $d_i$ is the rank difference of the $i$-th test image in the ground truth and predicted scores, it is given by:
\begin{equation}
    \operatorname{SRCC} =1-\frac{6\sum^{N}_{i=1}d^2_i}{N(N^2-1)},
\end{equation}
Both PLCC and SRCC provide insights into the model's performance, with higher values indicating better accuracy and consistency with the ground truth.

\subsection{Training Details}
\noindent\textbf{Degradation Encoder.}
The Degradation Encoder is tailored to extract and encode degradation features inherent in the input face images. Our architecture employs a CNN comprising six $3\times3$ convolution blocks. Each block incorporates batch normalization and is succeeded by a leaky ReLU activation. After feature extraction, these features are processed through a two-layer MLP to produce the final degradation representation vector.
We use the Adam optimizer with a learning rate of $3\times10^{-5}$ across 300 epochs for training. 
Our training data is divided into two distinct sets. The first set, labeled as \textit{Set $\mathcal{S}$}, consists of \( m \) images, as mentioned in Section 3.2 of the main paper. 
These are derived from 5000 high-quality images from the FFHQ dataset~\cite{karras2019style}, resized to $512\times512$. 
The images in this set are subjected to 15 different synthesized degradations, while one image remains undegraded, resulting in a total of 16 images (\textit{i.e.,} \(m=16\)). 
The synthesized degradations encompass a variety of conditions such as low-light, high-light, blur, defocus, 2x downsample, Gaussian noise, Gaussian blur with kernel sizes from 3 to 31, JPEG compression quality ranging from 1 to 30, motion blur, sun flare, ISO noise, shadow, and zoom blur. 
The low-light and high-light degradations are implemented using the torchvision library, whereas the other degradations are applied using albumentations~\cite{info11020125}.
The second set, designated as \textit{Set $\mathcal{R}$}, includes $n$ images, amounting to 256 as specified in Section 3.2.2 of the main paper. This set is dynamically curated by selecting from the GFIQA-20k dataset, ensuring that each subset of 256 images contains at least one high-quality face image with a Mean Opinion Score (MOS) greater than 0.9. The temperature parameter $\theta$ is 1.0. Notably, both sets undergo resampling in each iteration to ensure a diverse training experience. This module comprises a total of $1.27\times10^{6}$ parameters. The training process was conducted on a single NVIDIA A100 GPU, equipped with 80GB of memory, using the PyTorch framework. The entire training was completed in roughly 12 hours.
\noindent\smallskip\\
\noindent\textbf{Landmark Detection Network.}
We used a commercial implementation of \cite{deng2020retinaface} which outputs 1313 landmarks by fitting
the 3DMM model \cite{egger20203d} on the initially detected 68 landmarks.
We have observed that the original face landmark detection algorithm does not perform well on low-quality images. However, when fine-tuned specifically for low-quality images, it significantly improves performance, as shown in \figref{fig:landmark}.
These low-quality images are synthesized based on the image degradation model~\cite{Jo_2023_WACV} on the current landmark detection dataset.

\begin{figure}[t!]
\centering \includegraphics[width=0.48\textwidth,page=1]{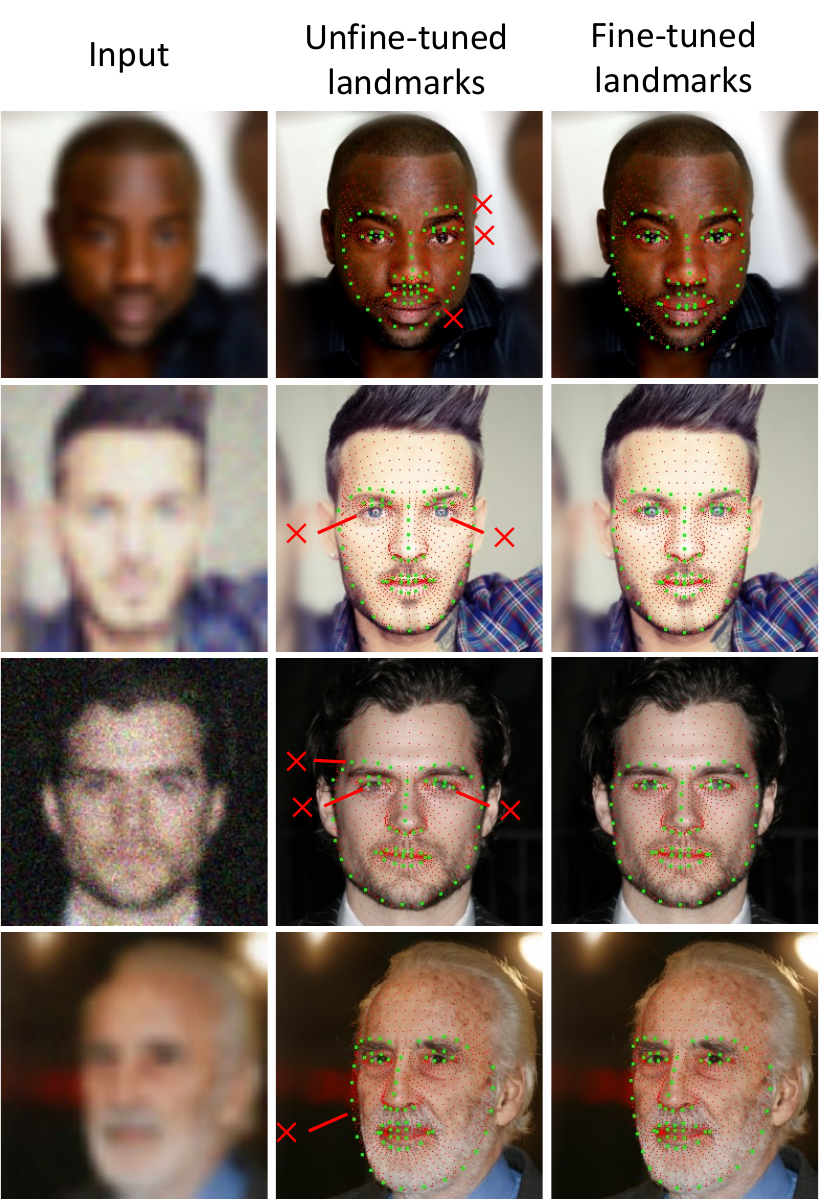}{}
\makeatother 
\caption{\textbf{Evaluating the Impact of Fine-Tuning on Landmark Detection in Poor-Quality Images.} The fine-tuned landmark detection algorithm can handle low-quality inputs (first column), as demonstrated in the third column of results. In contrast, the unfine-tuned algorithm has large errors, as evidenced in the second column (highlighted by the red crosses). The detected landmarks have been overlaid on the high-quality version of the input for better visualization. The basic 68 landmarks are represented by green dots, while the expanded set of 1313 landmarks is denoted by small red dots.} 
\label{fig:landmark}
\end{figure}
\noindent\smallskip\\
\noindent\textbf{GFIQA Network.}
The GFIQA Network, informed by the features extracted by the Degradation Encoder, endeavors to predict the Mean Opinion Score (MOS) for input face images. 
Our network architecture combines a hybrid CNN-Transformer backbone, comprising a VGG-19 model pre-trained on ImageNet~\cite{li2010crowdsourcing}, and a Vision Transformer (ViT) backbone~\cite{dosovitskiy2020image}, also pre-trained on ImageNet. This setup is further enhanced with two Swin Transformer blocks~\cite{liu2021swin}, a channel attention layer~\cite{hu2018squeeze}, a transformer decoder, and two MLP regression layers. The ViT backbone, tailored for an input size of $384\times384$, processes the image by dividing it into multiple $16\times16$ pixel patches, ensuring detailed and comprehensive image analysis.
During training, we employ a batch size of 16, and all input images undergo random cropping from $512\times512$ to $384\times384$. Additionally, data augmentation in the form of random horizontal flipping is applied to enhance the model's generalization capability. The learning rate is set at $10^{-5}$ across 100 epochs, and we use the Adam optimizer. The $\epsilon$ in $\mathcal{L}_{char}$ is $10^{-3}$. The module consists of $2.51\times10^{8}$ parameters in total. The network was trained on an Nvidia A100 GPU, which has 80GB of memory, using the PyTorch framework. The entire training process was completed within 20 hours.
\noindent\smallskip\\
\noindent\textbf{Clarification.} To clarify, in our system, both the degradation extraction network and the landmark detection network process the entire image ($512\times512$ pixels) to predict landmarks and extract degradation representations. However, for the GFIQA network, we adapt to the input size requirements of the pre-trained Vision Transformer (ViT), which is $384\times384$ pixels in our implementation. To accommodate this, we crop the facial image into several overlapping $384\times384$ patches, each serving as an individual input for the ViT. This ensures that the total coverage area of all patches encompasses the original input image.

In the main paper, particularly in Fig. 2, we simplified the explanation by omitting the step of cropping the facial image into multiple patches. Moreover, the images outlined in red in the GFIQA Network section are intended to illustrate how the ViT divides the input image ($384\times384$) into several patches for feature extraction between patches.

\section{More Experimental Results}
\subsection{Cross-Dataset Validation}
To explore the quality attributes of facial data, we conducted an experiment using our newly proposed CGFIQA-40k dataset and the existing GFIQA-20k~\cite{su2023going} dataset to train models. In this experiment, we employed the StyleGAN-IQA model~\cite{su2023going} and our method for training. The effectiveness of these models was then verified on the PIQ23 dataset~\cite{Chahine_2023_CVPR}, a benchmark for unseen face image quality assessment.

As shown in \tabref{tab:cross}, we observed that models trained on our datasets, particularly the CGFIQA-40k, demonstrated superior performance on the PIQ23 dataset\footnote{We test on device-exposure subset in PIQ23 dataset.}, an unseen face image quality dataset. This enhanced performance can be attributed to several key factors. Firstly, the CGFIQA-40k dataset is extensive in scale, encompassing a wide range of image qualities and scenarios. Secondly, and crucially, it offers a more balanced representation in terms of gender and skin tone compared to the GFIQA-20k dataset. This balanced representation ensures a more comprehensive and unbiased training process, leading to models that are better equipped to handle a diverse array of facial images in real-world applications. The results clearly highlight the advantages of our dataset, underscoring its potential in advancing the field of facial image quality assessment.

\begin{table}[t!]
\centering
\scalebox{0.8}{
\begin{tabular}{ccc} 
\toprule
\textbf{Dataset/Model }      & PLCC & SRCC \\ 
\hline\hline
GFIQA-20k/StyleGAN-IQA &    0.3323    &   0.3421         \\
CGFIQA-40k/StyleGAN-IQA &   0.3541     &  0.3643          \\
GFIQA-20k/Ours &      0.3947  & 0.4165           \\
CGFIQA-40k/Ours &      \textbf{0.4229}  & \textbf{0.4653}           \\
\bottomrule
\end{tabular}}
\\
\caption{\textbf{Performance Comparison of Zero-shot GFIQA on PIQ23~\cite{Chahine_2023_CVPR} Dataset.} This table compares the effectiveness of models trained on CGFIQA-40k and GFIQA-20k datasets. The results highlight the superior performance of models using CGFIQA-40k, underscoring its larger scale and balanced diversity in gender and skin tones.}
\label{tab:cross}
\end{table}

\begin{table}[t!]
\centering
\scalebox{0.95}{
\begin{tabular}{cccc} 
\toprule
\textbf{Strategy} & Naive & Patch-based & DSL \\ 
\hline\hline
mAP &  39.21   & 52.30  & \textbf{72.1}           \\
\bottomrule
\end{tabular}}
\\
\caption{\textbf{Comparative degradation retrieval accuracy using DSL, patch-based, naive methods under real-world degradations, quantified by mAP scores.}}
\label{tab:ablation-DE-naive}
\end{table}

\subsection{More Ablation Studies}
\noindent\textbf{Effectiveness of DSL.}
In our experiments, we compared two approaches to validate the effectiveness of our dual-set design in contrastive learning. The first approach, which we refer to as the ``Naive method'', involves training a model exclusively on the synthetic set (\textit{Set $\mathcal{S}$}). In this method, positive pairs are formed from images with identical synthetic degradations, while negative pairs are composed of images with different degradations. This approach, however, showed limitations in generalizing to real-world images due to its sole reliance on synthetic degradations.

In contrast, our dual-set model integrates both synthetic (\textit{Set $\mathcal{S}$}) and real-world (\textit{Set $\mathcal{R}$}) degradations. This model is trained to recognize and adapt to a broader range of degradation patterns, encompassing both controlled synthetic and naturally occurring real-world degradations. As a result, it demonstrated superior generalization capabilities, particularly in diverse real-world scenarios. The comparative performance of these two approaches is detailed in \tabref{tab:ablation-DE-naive}, highlighting the significant advantage of our dual-set approach in achieving more effective generalization in extracting degradation representation.
\noindent\smallskip\\
\noindent\textbf{Effectiveness of Channel Attention.}
By integrating a channel attention block, our method achieves a more precise feature focus, enhancing face quality assessment. This improvement leverages the well-documented advantages of attention mechanisms within the domain of image analysis, effectively emphasizing crucial channels. The comparative results, demonstrating the impact of incorporating channel attention into our approach, are detailed in \tabref{tab:ablation-CA}.
\noindent\smallskip\\
\noindent\textbf{Effectiveness of Landmark Guidance.}
We examine the impact of landmark guidance by conducting an experiment in which we omit the landmark detection component from DSL-FIQA. We then assess the performance on the GFIQA-20k~\cite{su2023going} and CGFIQA-40k datasets, with the results detailed in \tabref{tab:ablation-landmark}. This evaluation demonstrates that incorporating landmark guidance improves the effectiveness of our method.

\begin{table}[t!]
\centering
\scalebox{0.8}{
\begin{tabular}{ccc} 
\toprule
GFIQA-20k     & w/o CA & w CA   \\ 
\hline\hline
PLCC/SRCC &   0.9738/0.9733   &   \textbf{0.9745}/\textbf{0.9740}   \\
\bottomrule
\end{tabular}}
\\
\caption{\textbf{Impact of Channel Attention on Model Performance.}}
\label{tab:ablation-CA}
\end{table}

\begin{table}[t!]
\centering
\scalebox{0.6}{

\begin{tabular}{ccccc} 
\toprule
 PLCC/SRCC    & StyleGAN-IQA& MANIQA & DSL-FIQA w/o landmark & DSL-FIQA \\ 
\hline\hline
GFIQA-20k  &   0.9673/0.9684  &  0.9614/0.9604 & \underline{0.9725}/\underline{0.9720} & \textbf{0.9745}/\textbf{0.9740}      \\
CGFIQA-40k &   0.9822/0.9821  &  0.9805/0.9809 & \underline{0.9855}/\underline{0.9852} & \textbf{0.9873}/\textbf{0.9880}      \\
\bottomrule
\end{tabular}}
\\
\caption{\textbf{Impact of Landmark Guidance on Model Performance.}}
\label{tab:ablation-landmark}
\end{table}
\section{Discussion}
In our Dual-Set Contrastive Learning (DSL) framework, we utilize the real-world image set (\(\mathcal{R}\)) to establish soft proximity mapping through the synthetic image set (\(\mathcal{S}\)). Theoretically, it is possible for two or more images in set \(\mathcal{R}\) to have identical degradation representations.

However, it is important to note that the likelihood of this occurrence is extremely low due to the complex and variable nature of image degradation in real-world scenarios. In practice, even degradations that appear visually similar can have distinct characteristics influenced by various factors such as environmental conditions, lighting, and camera settings. Therefore, while the theoretical possibility of identical degradation representations in two images exists, it is practically negligible.

Additionally, we examine the t-SNE results presented in Figure 4 of the main paper. Initially, we observe that Gaussian noise, which is random and impacts the entire image, fundamentally contrasts with blurs and compressions that specifically affect image structure. This distinction likely causes Gaussian noise to appear separate from other degradations in t-SNE visualizations. Furthermore, JPEG compression and low resolution both lead to a loss of image detail, with the former eliminating high-frequency information and the latter decreasing the pixel count. This commonality in their impact on image clarity might result in similar patterns within the t-SNE visualizations.

}

{\small
\bibliographystyle{ieee_fullname}
\bibliography{egbib}
}

\end{document}